\documentstyle[12pt]{article}

\begin{document}
\bibliographystyle{abbrv}

\title{
Treating Coordination with Datalog Grammars
}

\author{
 Veronica Dahl\\
 {\normalsize Logic Programming Group and} \\
 {\normalsize Computing Sciences Department}\ \
 {\normalsize Simon Fraser University} \\
 {\normalsize \em veronica@cs.sfu.ca} \\
 \and
 Paul Tarau\\
 {\normalsize D\'epartement d'Informatique} \\
 {\normalsize Universit\'e de Moncton } \\
 {\normalsize \em tarau@info.umoncton.ca}
 \and
 Lidia Moreno\\
 {\normalsize Depto. de Sistemas Inform\'aticos y Computaci\'on}\\
 {\normalsize Universidad Politecnica de Valencia} \\
 {\normalsize \em lmoreno@dsic.upv.es} \\
 \and
 Manolo Palomar\\
 {\normalsize Depto. de Tecnolog\'ia Inform\'atica y Computaci\'on} \\
 {\normalsize Universidad de Alicante}   \\
 {\normalsize \em mpalomar@dtic.ua.es} \\
}

\date{}

\maketitle

\begin{abstract}
In \cite{DahlGULP94} we studied a new type of DCGs,
Datalog grammars, which are inspired on database theory. Their efficiency
was shown to be better than that of their DCG counterparts under
(terminating) OLDT-resolution.  In this article we motivate a variant of
 Datalog grammars
which allows us a meta-grammatical treatment of coordination. This treatment
improves in some respects over previous work on coordination in logic
grammars, although more research is needed for testing it in other respects.

\end{abstract}


\section{Introduction}
In their most common implementations, logic grammars resort to list
representations of the strings being analyzed or synthesized.
This list-based implementation results in several deficiencies in logic
grammars, while other deficiencies are inherited from Prolog.  Datalog
grammars were born in order to address all these deficiencies, namely: an
infinite Herbrand Universe, non-termination, unnecessary recomputation,
structure creation on the heap, bottleneck for multi-threaded execution due to
the use of (sequential)
list data structures, and inability to work directly on files.

 In Datalog grammars, a given CF grammar is automatically translated into an
 assertional representation, first proposed by Robert Kowalski,
which is largely equivalent to the list-based one but which ensures, under
appropriate evaluation mechanisms such as OLDT resolution, that the
termination and complexity properties of the original CF-grammar are
preserved.  We have moreover shown that,
in restricted but useful cases, this can be achieved even in the
presence of extra arguments.

Coordination has long been a  difficult problem both in linguistics and in
language processing. The difficulty lies in that any two constituents
can be coordinated (even of different kind), and in that often some substring
that is explicit in one of the conjuncts is missing in the other. For instance,
Wood's example:
\vskip 10pt

{\em John drove the car through and completely demolished a window.}

{\flushleft exhibits}
a missing object ( ''a window") in the first conjunct, and a
missing subject (''John") in the second. Moreover, in representing these
coordinated sentences, say in some logical form, we must take care of not
requantifying "a window" when we reconstitute its meaning at the missing
point: the window driven through must be equated with the demolished one.

While humans have in general no trouble reconstituting these missing elements
and attaching the right semantics to them, it is a challenge to efficiently
spell out for a machine the regularities found in coordination
phenomena.

In this article we show how we can extend the
incremental evaluation implementation of Datalog grammars in order to
automatically extend a grammar which has no rules for coordination with a
meta-grammatical treatment which allows us to parse coordinated sentences.

Our treatment of coordination incorporates an adaptation of
recent work on ellipsis which
resorts to the idea of parallel structures
(\cite{Darlymple91,Asher93,Grover94}, but unlike these approaches
which stress semantic parallelism, we use both {\em syntactic} and
{\em semantic} parallelism.


\section{Background}

\subsection{Assertional Representation}

 In DLGs, a call to analyze "the martian disappeared",  for
instance, compiles into:

{\small \begin{verbatim}
'D'(the,0,1).      % "the" stretches between points 0 and 1"
'D'(martian,1,2).
'D'(disappeared,2,3).

?- sentence(0,3).  % Is there a sentence between points 0 and 3?
\end{verbatim}}

{\flushleft while}
lexical rules compile into forms that  use these representations
accordingly, e.g.:

{\small \begin{verbatim}

noun(P1,P2):- 'D'(martian,P1,P2).

\end{verbatim}}

{\flushleft Other} grammar rules translate just as in Definite Clause Grammars,
the standard Prolog grammar formalism.

\subsection{Incremental evaluation}

In order to increase efficiency, one possible implementation for for DLGs
exploits the incremental Datalog
technique of generating and maintaining data bottom-up. Using the well-known
semi-naive evaluation algorithm, we begin with the set of axioms and obtain the
theorems of the
first "layer" by applying the derivation rules; then we take these theorems
as new starting point, to derive the theorems of the second layer , and so
on. Generally to derive the theorems of the next "layer", at least one
theorem produced at the previous stage must be used. This process
terminates when no more new theorems can be generated.

\subsection{Coordination}

Early work on coordination proposed meta-grammatical treatments (e.g.
\cite{Woods73,DMcC83}), in which the appearance of a coordinating word, or
conjunction (e.g. ''and", ''or", ''but") is treated as a demon. When a
conjunction
appears in a
sentence of the form

 	{\center A X conj Y B}

{\flushleft a} process is triggered in which backing up is done in the parse
history in order
to parse Y parallel to X, and B is parsed by merger with the state interrupted
by the conjunction.

Thus, in Wood's example we would have:

{\small \begin{verbatim}
A= John
X= drove his car through
conj= and
Y= completely demolished
B= a window
\end{verbatim}}

The reconstructed phrase should then be A X B and A Y B, with the warning
already made re. requantification.

We next modify this treatment and express it through
DLG constraints to be intertwined with the incremental evaluation of a
DLG grammar.  We shall then discuss more recent views on parsing parallel
structures, and extend our treatment by adapting some of these ideas into our
DLG framework.

\section{Treating coordination through DLGs plus constraints}

Our idea for a Datalog treatment of coordination is also, as in the work
reviewed in the last section, based on the
assumption that a string containing a conjunction contains around that
conjunction two
constituents which are being coordinated. But instead of identifying
four substrings A, B, X and Y, we simply assume that there are two
coordinating constituents, V and W, surrounding the conjunction,
which must in general be of the same category and have parallel parses. Thus
any missing
elements in either V or W can be reconstructed from the other. We also adopt
the heuristics that closer scoped coordinations will be attempted before
larger scoped ones. Thus in Wood's example, ''vp conj vp" is tried before
''sent conj sent".

Thus in that example, ''John" would parse as the subject noun phrase of a
 sentence with
a complex verb phrase. Therefore we have

{\small \begin{verbatim}
V= drove his car through

W= completely demolished a window
\end{verbatim}}

Because the conjunction is reached before the first verb phrase is finished
parsing ("through" analyses as a preposition introducing a prepositional
phrase- i.e., expecting a noun phrase to follow), the unfulfilled expectation
of a noun phrase is postponed until it can be equated with a noun phrase in
W.

Notice that what we mean by V and W having parallel parses is not that they
must necessarily
follow the  same structure to the last details, but that their
structures must complement each other so that missing constituents in one may
be
reconstructed from the other. We further assume, for the purposes of this
article, that they both must
have the same root (in this case, a verb phrase root),  although this
assumption is not necessary in general.

Another thing to notice is that, whereas in the first analysis of Wood's
example we end up with two conjoined {\em sentences}, in the analysis just
proposed we end up with a sentence having a verb phrase which decomposes
into two conjoined verb phrases. Linguistically speaking, it is
arguable whether one analysis is preferable over
the other one. But computationally speaking, the second analysis allows us
to apply our meta-grammatical treatment of coordination to sentences for which
the first analysis would fail. An example is

{\em Jean mange une pomme rouge et une verte.}

 This sentence cannot be split into A X conj B Y to reconstitute an unreduced
structure following the first analysis. On the other hand, using the second
analysis, we can postulate

{\em V = une pomme rouge, W = une verte}

and require that W follow a structure parallel to that of V. This then allows
us to reconstitute the missing noun in W.

We now describe our proposed extension of incrementally implemented Datalog
grammars in an intuitive manner, using the above example. We assume a simple
French grammar with rules such as

{\small \begin{verbatim}
sent --> np, vp.                np --> proper_noun.

vp --> v, np.                   np --> det, n, adj.
\end{verbatim}}

Our grammar includes no rules for coordination (but does, of course,
recognize conjunctions as such).

Let us recall that, in a Datalog grammar, our input string would be
represented as:

{\small \begin{verbatim}
0 Jean 1 mange 2 une 3 pomme 4 rouge 5 et 6 une 7 verte 8.
\end{verbatim}}

The idea is simply to check, at every step of the incremental derivation
of the theorems, whether a theorem conj(N,M) has been derived. As soon as
one is, a constraint is added to the effect that, in some subsequent step
of the incremental derivation of theorems, a constituent of category Cat must
be found between some point Z and the point N, such
that the same category stretches between M and some later point P; and that
finding them implies that the string between Z and P must also have category
Cat.

This constraint can be noted:

{\small \begin{verbatim}
C= {Cat(Z,N) isomorphic with Cat(M,P) ==> add Cat(Z,P)}
\end{verbatim}}

As soon as one of these predictions is fulfilled (e.g. when we have found a
noun
phrase ''une pomme rouge" between Z=2 and N=5), we can further specify the
other prediction to follow the same structure as that of the found noun
phrase, which will allow us to reconstruct any missing elements.

Notice that backtracking can occur\footnote{This means, in case of Prolog's
usual execution strategy that something like linear implication
\cite{jlp43}
should be used instead of asserting to the dynamic database.
Even more appropriate for this purpose is BinProlog's
linear assumption operator assumel/1 \cite{Tarau95:BinProlog},
which ensures that
the assumed facts scope will range over the `continuation' i.e. it will
be true in all future computations belonging the same
resolution branch.}.
For instance, the machine will first
postulate that the conjoined categories must be ''adjective", and that Z=4
(this
would be a good guess for the sentence: ''Jean mange une pomme rouge et
verte").
But in our sample sentence, this first try will fail to find an adjective
starting at point M=6, so backtracking would undo the bindings and suspend the
constraint until
other suitable candidates for ''Cat" and ''Z" are derived.

We next present a step-by-step follow up for the example given. Sequences of
theorems derived are noted T1, T2, etc.; whereas sets of constraints are noted
C1,C2,etc.

{\small \begin{verbatim}
T1= {'D'(jean,0,1),...,'D'(rouge,7,8)}.

T2= T1 union {n(0,1), v(1,2), det(2,3), n(3,4), adj(4,5), conj(5,6),
det(6,7), adj(7,8)}
\end{verbatim}}

{\small \begin{verbatim}
C2 = { Cat(Z,5) isomorphic with Cat(6,P) ==> add Cat(Z,P)}
\end{verbatim}}

{\flushleft tries} Z=4 and fails. So the
constraint suspends until something else of the form Cat(Z,5) appears.

{\small \begin{verbatim}
T3 = T2 union {np(2,5), np(0,1)}
\end{verbatim}}

{\flushleft tries} Z=2, and uses top-down prediction to find a
(possibly incomplete) noun phrase stretching from point 6 to some
point P, e.g. through the rule:

{\small \begin{verbatim}
np(6,P):- det(6,X), n(X,Y),adj(Y,P).
\end{verbatim}}

{\flushleft succeeds} with  substitutions {X=7, Y=7,P=8}

{\small \begin{verbatim}
T4 = T3 union {np(6,8)} union {np(2,8)}

T5 = T4 union {vp(1,8)}

T6 = T5 union {s(0,8)}
\end{verbatim}}

Notice that, at the point in which the constraint succeeds with  substitutions
{X=7, Y=7}, if the grammar included arguments for semantic representation,
the semantic representations for the two nouns would be unified, given that
one of them is missing (as shown by the fact that its starting point, 7, is the
same as its ending point). We shall later give a full example involving
semantic representations.

\section{Related work on ellipsis}

A notion that is central to recent work on ellipsis, and which has been
present in embryonic form, as we have seen, even in the early work on
coordination, is that of parallelism as a key element in the determination of
implicit meanings. Asher \cite{Asher93} defines parallelism as

\begin{quote}
a pairing of constituents ... and their parts, such that each pair contains
two semantically and structurally similar objects
\end{quote}

{\flushleft \cite{Darlymple91} describes}
 an elliptical construction as
\begin{quote}
one involving two phases
(usually clauses) that are parallel in structure in some sense.
\end{quote}

{\flushleft \cite{Grover94}}, following \cite{Prust92}, also postulates the
necessity,
within a feature-structure setting, of combining elements which exhibit
a degree of syntactic-semantic parallelism in order to determine the
way in which some kinds of anaphora are resolved, and argue that the use
of default unification (or priority union) improves on
Pr\"ust's operation for
combining the parallel structures.

Although the analysis of \cite{Darlymple91} precedes that of \cite{Grover94},
the latter may be easier to follow, so we shall discuss it first.

Intuitively, default unification \cite{Calder90} takes two feature
structures, one of which (called the TARGET) is identified as ''strict",
while the other one (called the SOURCE) is ''defeasible", and combines
the information in both such that the information in the strict structure
takes priority over that in the defeasible structure. For instance,
the combination of the feature structures shown below for sentences 1a and
1b

{\small \begin{verbatim}
1a. Hannah likes beetles.

        [ AGENT Hannah
          PATIENT beetle]
        likes

1b. So does Thomas

        [ AGENT Thomas]
        agentive
\end{verbatim}}

{\flushleft results} in the priority union:

{\small \begin{verbatim}

         [ AGENT Thomas
           PATIENT beetle]
         likes
\end{verbatim}}


Thus, the implicit constituent in the second sentence is reconstituted from the
first by using a generally applicable procedure on the representations of the
parallel structures.

\cite{Darlymple91} postulated a similar analysis, but it was based on
$\lambda$-calculus semantic representations, and used higher order unification.

For instance, in their example:
\vskip 10pt
{\em Dan likes golf, and George does too.}

{\flushleft they} identify the antecedent or source as the complete structure
(''Dan likes
golf"), whereas the target clause (''George does too") is either missing,
or contains only vestiges of, material found overtly in the source.

Their analysis of such structures consists of:

{\flushleft a)} determining the parallel structure of source and target;

{\flushleft b)} determining which are parallel elements in source and target
(e.g.,
''Dan" and ''George" are parallel elements in the example);

{\flushleft c)} using Huet's higher-order unification algorithm \cite{Huet75}
for finding
a property P such that $P(s_1,...,s_n)=S$,

{\flushleft where} $s_1$ through $s_n$ are the
interpretations of the parallel elements of the
source, and s is the interpretation of the source itself. Only solutions
which do not contain a primary occurrence of the parallel elements are
considered (occurrences are primary if they arise directly from the parallel
elements, as opposed to those arising for instance from a pronoun).

{\flushleft In the example},

{\small \begin{verbatim}
	P(dan) = likes(dan,golf)
\end{verbatim}}

{\flushleft is} solved by equating P with

$\lambda$x. likes(x,golf)

{\flushleft given} that the other possible solution,
$\lambda$x. likes(dan,golf) contains
a primary occurrence of the parallel element, ''dan", and must therefore
be discarded.

{\flushleft d)} applying the property on the representation of the target, e.g.

   P(george)= [$\lambda$x.likes(x,golf)] george = likes(george,golf)

{\flushleft e)} conjoining the meanings of the source and of the target thus
completed,
e.g.:

{\small \begin{verbatim}
	likes(dan,golf) & likes(george,golf)
\end{verbatim}}

{\flushleft Both} \cite{Darlymple91}
and  \cite{Grover94} provide ambiguous
readings of discourses such as

{\small \begin{verbatim}
	Jessy likes her brother. So does Hannah.
\end{verbatim}}

{\flushleft can} be provided, unlike previous analyses,
 without having to postulate ambiguity in the source (this is achieved in
\cite{Grover94} by allowing for priority union to either
preserve or not preserve structure-sharing information in the source, and in
 \cite{Darlymple91} by the distinction between primary and secondary
occurrences
of parallel elements). Another notable point    in both
these approaches is that they address the issue of semantic parallelism, which
in most previous approaches was understressed in favor of syntactic
parallelism.

However, both methods share the following limitations:

{\flushleft a)} neither method formulates exactly how parallelism is to be
determined- it
is just postulated as a prerequisite to the resolution of ellipsis (although
\cite{Grover94} speculates on possible ways of formulating this, leaving it
for future
work)

{\flushleft b)} both approaches stress semantic parallelism, while pointing out
that
this is not sufficient in all cases

By examining ellipsis in the context of coordinated structures, which are
parallel by definition, and by using extended DLGs, we provide a method in
which parallel structures are detected and resolved through syntactic and
semantic criteria, and which can be applied to either grammars using different
semantic representations- feature structure, $\lambda$-calculus, or other. We
exemplify using a logic based semantics along the lines of \cite{DMcC83}.

\subsection{Our semantico-syntactic treatment of parallelism}

{\flushleft Let} us now consider the  string

\begin{verbatim}
 John drove the car through and demolished a window
0    1     2   3   4       5    6         7 8      9
\end{verbatim}

{\flushleft where} we
have indicated the connections as numbers in between the words.

We  use the following grammar:


{\small \begin{verbatim}
sent(Sem) --> np(X,Scope,Sem), vp(X,Scope).

np(X,Scope,Sem) --> det(X,Restriction,Scope,Sem), noun(X,Restriction).
np(X,Sem,Sem) --> name(X).

vp(X,Sem) --> verb0(X,Sem).
vp(X,Sem) --> verb1(X,Y,S0), np(Y,S0,Sem).
vp(X,Sem) --> verb2(X,Y,Z,S0), np(Y,S0,S1), pp(Z,S1,Sem) .

pp(X,S0,Sem) --> prep(_), np(X,S0,Sem).

verb0(X,laugh(X)) --> [laughed].
verb1(X,Y,ate(X,Y)) --> [ate].
verb1(X,Y,saw(X,Y)) --> [saw].
verb1(X,Y,heard(X,Y)) --> [heard].
verb1(X,Y,demolished(X,Y)) --> [demolished].
verb2(X,Y,Z,drove_through(X,Y,Z)) --> [drove].
verb1(X,Y,sat_at(X,Y)) --> [sat].

det(X,Scope,Restriction,exists(X,Scope,Restriction)) --> [a].
det(X,Scope,Restriction,exists(X,Scope,Restriction)) --> [an].
det(X,Scope,Restriction,def(X,Scope,Restriction)) --> [the].
det(X,Scope,Restriction,each(X,Scope,Restriction)) --> [each].

noun(X,man(X)) --> [man].
noun(X,woman(X)) --> [woman].
noun(X,apple(X)) --> [apple].
noun(X,pear(X)) --> [pear].
noun(X,window(X)) --> [window].
noun(X,table(X)) --> [table].
noun(X,train(X)) --> [train].
noun(X,car(X)) --> [car].

name(john) --> [john].
name(mary) --> [mary].

prep(through) --> [through].
prep(at) --> [at].

conj(and) --> [and].
\end{verbatim}}

{\flushleft T1} would contain the 'D' connections for this sentence, and T2
adds:

\begin{verbatim}
{name(john,0,1), verb2(X,Y,Z, drove_through(X,Y,Z),1,2),
  det(Y,R,Sc,the(Y,R,Sc),2,3), noun(Y,car(Y),3,4),
  prep(through,4,5), conj(and,5,6), verb1(X1,Y1,
  demolished(X1,Y1),6,7), det(W,R1,Sc1,a(W,R1,Sc1),7,8),
  noun(V,window(V),8,9)}
\end{verbatim}

{\flushleft At} this point, a constraint to find points P and Q such that
{\tt Cat(...,P,5)}
is parallel to {\tt Cat(...,6,Q)} is generated (upon whose finding, something
of the form
{\tt Cat(..., P,Q)} will be added to the set of theorems~\footnote{We shall
describe below the constraints that this addition must satisfy.}), and this
constraint
 suspends until the following new theorems have been derived:

{\small \begin{verbatim}
T4= T2 union {np(john,Sem,Sem), np(Y,Sc,the(Y,car(Y),Sc),2,4),
    np(W,Sc1,a(W,window(W),Sc1),7,9), vp(X,a(W,window(W),
    demolished(X,W),6,9)}

\end{verbatim}}

{\flushleft We} can now postulate Cat= vp and use top-down prediction to derive
a
(possibly incomplete) vp ending at point 5. When trying rule

{\small \begin{verbatim}
vp(X,Sem) --> verb2(X,Y,Z,S0), np(Y,S0,S1), pp(Z,S1,Sem).

\end{verbatim}}

{\flushleft and} after development of the pp, we can adapt the analysis of
\cite{Darlymple91} and identify:

{\small \begin{verbatim}
SOURCE NP:

np(W,demolished(X,W),a(W,window(W),demolished(X,W)),2,4)

TARGET NP:

np(Z,the(Y,car(Y),drove_through(X,Y,Z),Sem',5,5)

\end{verbatim}}

{\flushleft Now} we can build an abstract NP by abstracting over the Scope
argument of the source, and postulating an empty surface string (i.e., by
equating the start
and end points of the string):

{\small \begin{verbatim}
ABSTRACTED NP:

np(W,Scope,a(W,window(W),Scope),P,P)

\end{verbatim}}

{\flushleft We} now unify the abstracted NP with the target NP to obtain the
resolved
target NP:

{\small \begin{verbatim}
RESOLVED TARGET NP:

np(W,the(Y,car(Y),drove_through(X,Y,W),a(W,window(W),the(Y,car(Y),
drove_through(X,Y,W),5,5)

\end{verbatim}}

{\flushleft which} in turn completes the target vp:

{\small \begin{verbatim}
RESOLVED TARGET VP:

vp(X',a(W,window(W),the(Y,car(Y),drove_through(X,Y,W))),1,5)
\end{verbatim}}

{\flushleft The} constraint now reads:

{\small \begin{verbatim}
C2'= {vp(X',a(W,window(W),the(Y,car(Y),drove_through(X,Y,W),1,5)
	parallel to
      vp(X,a(V,window(V),demolished(X,V),6,9) }
\end{verbatim}}

{\flushleft Now} we
need to conjoin the parallel structures. This is done by what we
call \verb~c-unification~: unify the
parts in the parallel terms which are unifiable, and conjoin  those that
are not(i.e., the parallel elements), with the exception of the last two
arguments, which are generating from the two pairs of last arguments P1-P2 and
P2+1-P3 of the parallel structures, as P1 and P3~\footnote{Of course, for other
types of coordination we use the appropriate connective: ''but", ''or",...}. We
obtain:

{\small \begin{verbatim}
vp(X,a(W,window(W),and(the(Y,car(Y),drove_through(X,Y,W)),
    demolished(X,W)),1,9)
\end{verbatim}}

After this theorem's addition, the sent rule can apply to derive

{\small \begin{verbatim}
sent(a((W,window(W),and(the(Y,car(Y),drove_through(john,Y,W)),
    demolished(john,W)),1,9)
\end{verbatim}}

\section{Discussion}

The previous section shows that when we introduce syntactic as well as
semantic parallelism, this can help determine which are the parallel
structures automatically, by incremental application of a Datalog grammar
constraint on coordination coupled with top-down prediction to complete
any missing structures through an analysis of parallelism that is inspired in
that of \cite{Darlymple91} but complements it in various ways. Syntactic
criteria on the determination of parallelism that can be found in the
literature can also, of course, be added to complement this initial proposal.

Several observations are in order. In the first place, we must note that a
simple conjoining of the representations obtained for the parallel
structures as proposed in \cite{Darlymple91} may not, as  the example of the
previous section shows, suffice. Since these structures are quite dissimilar,
we must conjoin only the parallel elements. We  postulate that the parallel
elements will be represented by those subterms which are not unifiable.

Secondly, our notion of   abstraction, which relies on converting into a
variable those parts of a semantic representation which are contributed by the
constituent that contains it, can be adapted to suit
other semantic representations, provided that we can identify for them
which part of the semantic representation each rule for a constituent
contributes to the overall representation. This is not an unreasonable
expectation for compositionally defined semantics.

In the third place, we should note that our analysis allows for the source
clause to not necessarily be the first one- again as the example we just
examined shows, we can have structures in which the incomplete substructure
does not  antecede  the complete one. Thus our analysis can handle more
cases than those in previous related work.

Note that some special cases allow to use unification
between isomorphical objects to obtain the proper quantification.
By slightly modifying the grammar as

{\small \begin{verbatim}
np(X,Scope,Sem) -->
   np0(X,Scope,Sem). % previously defined as np/3

np(X,Scope,and(Sem1,Sem2)) -->
   np0(X,Scope,Sem1),
   conj(and),
   np(X,Scope,Sem2).

\end{verbatim}}

{\flushleft we} can handle directly phrases like:

{\small \begin{verbatim}

                    Each man ate an apple and a pear.

                                   ||
                                   \/

                                  each
                                    |
               .---.--------------------.
               |   |                    |
              V0  man                  and
                   |                    |
                   .           .------------------.
                   |           |                  |
                  V0        exists             exists
                               |                  |
                        .----.-------.     .----.------.
                        |    |       |     |    |      |
                       V1  apple    ate   V1  pear    ate
                             |       |          |      |
                             .     .---.        .    .---.
                             |     |   |        |    |   |
                            V1    V0  V1       V1   V0  V1
\end{verbatim}}

{\flushleft Clearly} this
works only for a class of particular constraints exhibiting strong isomorphism
in the constructed meaning.
For instance, noun groups of the form $np_1$ {\tt and} $np_2$ {\tt and} $np_3$
do have this property.

We must note, however, that in some cases we will need to complement our
analysis with a further phase which we shall call ''reshaping". Take for
instance the sentence ''Each man and each woman ate an apple". Since both
parallel structures are complete, we do not need to perform abstraction
and \verb~c-unification~, but we do need to reshape the result of the analysis
through distribution, thus converting

{\small \begin{verbatim}
each(X,man(X)&woman(X),exists(Z,apple(Z),ate(X,Z))
\end{verbatim}}

{\flushleft into}

{\small \begin{verbatim}
and(each(X,man(X),exists(Z,apple(Z),ate(X,Z)),
   each(X,woman(X),exists(Z,apple(Z),ate(X,Z)))
\end{verbatim}}

Reshaping operations have been used in \cite{DMcC83}, and are useful in
particular to decide on appropriate quantifier scopings where coordination is
involved. It would be interesting to study how to adapt these operations
to the present work.

Another interesting observation is that the results in \cite{Darlymple91}
concerning the use of the distinction between primary and secondary occurrences
of parallel elements in order to provide ambiguous readings of discourses
 such as
''Jessie likes her brother. So does Hannah." could in principle be transferred
into our approach as well.

Let us also note that, as observed in \cite{Asher93}, the notion of
compositional
semantics of the two clauses (on which the related previous work, and ours
to some extent, is based) is not enough in some cases. For instance, consider:

{\small \begin{verbatim}
   If Fred drinks, half the bottle is gone.
   But if Sam drinks too, the bottle is empty.
\end{verbatim}}

In this sentence, the conclusion which holds if Fred drinks BUT SAM DOES NOT,
does not hold if both Fred and Sam drink. The implicit information that
the first conclusion holds only if the premiss of the second sentence does not
hold must be inferred. Using our approach, we could use the re-shaping
phase to deal with cases such as this one, in which the presence of
words such as ''too" would trigger the generation of the full reading.
A sentence of the form

{\small \begin{verbatim}
   If Fred drinks, C1, but if Sam drinks too, C2.
\end{verbatim}}

{\flushleft would} generate a representation such as

{\small \begin{verbatim}
   but(if(drink(fred),C1),if(too(drink(sam)),C2))
\end{verbatim}}

{\flushleft which} after reshaping would become:

{\small \begin{verbatim}
  and(if(and(drink(fred),no(drink(sam)),C1),
      if(and(drink(fred),drink(sam),C2)))
\end{verbatim}}

Finally, let us point out that, unlike most current efforts on programming
 with constraints,   the constraints we propose in this paper
do not limit themselves to pruning the search
space, but actively contribute to finding a solution. In this sense they
relate more to database work such as \cite{Sri93} than to the literature
in either constraint logic programming or constraint logic grammars.

\section*{Acknowledgement}
This research was supported by NSERC Research grants 31-611024 and OGP0107411,
and by NSERC, CSS and SFU PRG Infrastructure and Equipment grant given to the
Logic and Functional Programming Laboratory at SFU, in whose facilities
part of this work was developed. We are also grateful to the Centre for Systems
Science, LCCR and the School of Computing Sciences at Simon Fraser University
for the use of their facilities. Paul Tarau also thanks for support from the
FESR of the Universit\'{e} de Moncton.


\begin{thebibliography}{10}

\bibitem{Asher93}
N.~Asher.
\newblock {\em Reference to Abstract Objects in Discourse}, volume~50 of {\em
  Studies in Linguistics and Philosophy}.
\newblock Kluwer, 1992.

\bibitem{Calder90}
J.~H.~R. Calder.
\newblock {\em An Interpretation of Paradigmatic Morphology}.
\newblock PhD thesis, University of Edinburgh, 1990.

\bibitem{DMcC83}
V.~Dahl and M.~McCord.
\newblock Treating coordination in logic grammars.
\newblock {\em American Journal of Computational Linguistics}, 9:69--91, 1983.

\bibitem{DahlGULP94}
V.~Dahl, P.~Tarau, and Y.~N. Huang.
\newblock Datalog grammars.
\newblock In {\em Proc. 1994 Joint Conference on Declarative Programming},
  Peniscola, Spain, September 1994.

\bibitem{Darlymple91}
M.~Darlymple, S.~Shieber, and F.~Pereira.
\newblock Ellipsis and higher-order unification.
\newblock {\em Linguistics and Philosophy}, 14(4):399--452, 1991.

\bibitem{Grover94}
C.~Grover, C.~Brew, S.~Manandhar, and M.~Moens.
\newblock Priority union and generalization in discourse grammars.
\newblock In {\em Proc. 32nd ACL Conference}, New Mexico, 1994.

\bibitem{jlp43}
J.~Hodas.
\newblock Specifying {F}iller-{G}ap {D}ependency {P}arsers in a {L}inear
  {L}ogic {P}rogramming {L}anguage.
\newblock In K.~Apt, editor, {\em Proc. 1992 Joint International Conference and
  Symposium on Logic Programming}, pages 622--636. MIT Press, 1992.

\bibitem{Huet75}
G.~Huet.
\newblock A unification algorithm for typed lambda-calculus.
\newblock {\em Theoretical Computer Science}, 1:27--57, 1975.

\bibitem{Prust92}
H.~Prust.
\newblock {\em On Discourse Structuring, Verb Phrase Anaphora and Gapping}.
\newblock PhD thesis, Universiteit van Amsterdam, 1992.

\bibitem{Sri93}
D.~Srivastava and R.~Ramakrishnan.
\newblock Pushing {C}onstraint {S}elections.
\newblock {\em The Journal of Logic Programming}, 16:361--414, 1993.

\bibitem{Tarau95:BinProlog}
P.~Tarau.
\newblock Bin{P}rolog 3.30 {U}ser {G}uide.
\newblock Technical Report 95-1, D\'{e}partement d'Informatique, Universit\'{e}
  de Moncton, Feb. 1995.
\newblock ftp://clement.info.umoncton.ca/{B}in{P}rolog.

\bibitem{Woods73}
W.~Woods.
\newblock An experimental parsing system for transition network grammars.
\newblock In R.~Rustin, editor, {\em Natural Language Processing}, pages
  145--149. Algorithmic Press, New York, 1973.

\end{thebibliography}

\end{document}